%% file: main.tex
\documentclass{article}

\usepackage{microtype}
\usepackage{graphicx}
\usepackage{subfigure}
\usepackage{booktabs} 
\usepackage{dsfont}

\usepackage{amsmath}
\usepackage{amsfonts}
\usepackage{amssymb}
\usepackage{hyperref}

\usepackage[accepted]{icml2021}

\icmltitlerunning{GLAM: Graph Learning by Modeling Affinity}

\begin{document}

\twocolumn[
\icmltitle{GLAM: Graph Learning by Modeling Affinity to Labeled Nodes for Graph Neural Networks}

%



\icmlsetsymbol{equal}{*}

\begin{icmlauthorlist}
\icmlauthor{Vijay Lingam}{msr}
\icmlauthor{Arun Iyer}{msr}
\icmlauthor{Rahul Ragesh}{msr}
\end{icmlauthorlist}

\icmlaffiliation{msr}{Microsoft Research India}

\icmlcorrespondingauthor{Vijay Lingam}{vijaylingam0810@gmail.com}

\icmlkeywords{Machine Learning, ICML}

\vskip 0.3in
]



\printAffiliationsAndNotice{\icmlEqualContribution} 

\input{sections/abstract}
\input{sections/introduction}
\input{sections/problem}
\input{sections/related}

\input{sections/model}
\input{sections/experiment}
\input{sections/conclusion}

\bibliography{references}
\bibliographystyle{icml2021}

\end{document}

%% file: sections/abstract.tex
\begin{abstract}
Graph Neural Networks have shown excellent performance on semi-supervised classification tasks. However, they assume access to a graph that may not be often available in practice. In the absence of any graph, constructing \textit{k}-Nearest Neighbor (\textit{k}NN) graphs from the given data have shown to give improvements when used with GNNs over other semi-supervised methods. This paper proposes a semi-supervised graph learning method for cases when there are no graphs available. This method learns a graph as a convex combination of the unsupervised \textit{k}NN graph and a supervised label-affinity graph. The label-affinity graph directly captures all the nodes' label-affinity with the labeled nodes, i.e., how likely a node has the same label as the labeled nodes. This affinity measure contrasts with the \textit{k}NN graph where the metric measures closeness in the feature space. Our experiments suggest that this approach gives close to or better performance (up to 1.5\%), while being simpler and faster (up to 70x) to train, than state-of-the-art graph learning methods. We also conduct several experiments to highlight the importance of individual components and contrast them with state-of-the-art methods. 
\end{abstract}

%% file: sections/introduction.tex
\section{Introduction}
\label{sec:introduction}

Incorporating graph to improve semi-supervised classification is a long and well-studied problem. Early methods used graph as a means to propagate labels~\cite{lp_zhu, loc_glob_const}. Classifiers later used graphs for regularization~\cite{manifold_reg, embednn}. Recently, a new class of neural network called Graph Neural Networks (GNNs)~\cite{survey_wu} have been proposed, the most prominent among them being Graph Convolutional Network (GCN)~\cite{gcn}. These models can incorporate graphs into their architecture to aid the learning process. These models have shown performance exceeding approaches that only use the graph to propagate labels or to regularize. However, all these approaches assume the existence of a graph. In this paper, we are mainly interested in scenarios where we do not have access to a graph. In the absence of graphs, utilizing \textit{k}-Nearest Neighbor (\textit{k}NN) graphs have proven to be quite effective~\cite{manifold_reg, knn-gnn} and give improvements over models that do not use graph information. 

The GNN models incorporating graphs into their architecture have opened up the possibility of learning graphs from the training data in a semi-supervised fashion. Recent papers~\cite{lds, idgl} have proposed such graph learning algorithms. In~\cite{lds}, the authors assume access to an initial graph and propose a generative model over the graph. The entire optimization problem is cast as a bi-level programming problem where the inner objective is to learn GNNs on the training data. The outer objective is to learn the generative model parameters over the validation set. In~\cite{idgl}, the authors propose to take an initial graph and refine it using the latent node representations learnt by GNNs. The model uses this refined graph to learn better representations. This cycle repeats until the stopping condition is met. In the absence of any given graph, both these approaches utilize \textit{k}NN graphs. The training procedures for these models can be complex and expensive. Also, the quality of graphs constructed using \textit{k}NN can vary from dataset to dataset. Both these approaches learn graphs implicitly via the GNN model and thus are dependent on the quality of the graphs provided to them. Hence, our work's goal has been to propose a simpler model, which is faster to train while being robust to noisy graphs.

An alternative form of learning graphs is to use attention models. Graph Attention Network (GAT)~\cite{gat} proposed to learn attention over edges of a graph. The authors proposed GAT as an alternative to computing Laplacians for large graphs in GCN. However, it is not clear what these attention models in graphs tend to learn. For example, do they capture any meaningful relationship between the nodes? To address this issue, SuperGAT~\cite{supergat} proposed an improved version of this attention model. This improvement in attention comes by adding a self-supervised loss term that uses the existing edges in the graph as labeled data. Learning to model attention on all the edges of a graph given small labeled data is challenging. Adding the self-supervised loss does mitigate that problem to some extent. However, since \textit{k}NN graphs are quite noisy, any gains in performance may be nullified by the graph's noise. Also, attention models restrict themselves to learn attention over the edges present in the graph and may not see certain valuable long-range interactions needed, particularly in \textit{k}NN graphs. We seek inspiration from attention-based models and attempt to address some of the limitations.

We observe that existing approaches are either expensive to train, or rely on latent representations learned from GNN's, which can be poor when using noisy graphs. In this work, we propose a simpler and cheaper model for learning graphs that are as competitive or better than the state-of-the-art models in terms of performance for datasets whose \textit{k}NN graphs are very noisy. The key contributions of our work are:
\begin{enumerate}
    \item Inspired by the works in attention-based models, we propose a novel label-affinity model, which attempts to predict for any given input node, which node in the labeled set is likely to have the same label as the input node.
    \item Following the work of~\cite{idgl}, we propose to learn a convex combination of an unsupervised \textit{k}NN graph and the graph generated using the label-affinity model (supervised graph).
    \item Following the work of~\cite{supergat}, we also propose an explicit loss term for the label-affinity model; however, instead of using semi-supervised loss term, we use a supervised loss term.
    \item We show that combining these three ideas gives rise to a straightforward model that is faster (up to 70x) to train while giving close to or better (up to 1.5\%) improvement in performance.
\end{enumerate}
In the next few sections, we will discuss the problem formally, our proposed model, related work and results in more detail.

%% file: sections/problem.tex
\section{Problem Statement}
\label{sec:problem}

In the semi-supervised classification problem, we have labeled examples $D = \{(x_i, y_i)\}_{i=1}^l$ and unlabeled examples $U = \{x_i\}_{i=l+1}^{l+u}$, where $x_i \in X$. Let $Y$ be the set of all possible labels and $\mathbf{y}$ be the vector $[y_1, y_2, \ldots, y_{l+u}]$. The goal in traditional semi-supervised classification problem is to learn a function $f:X \mapsto Y$ and it is often learnt by solving the following minimization problem,
\begin{gather}
    f^* = \arg \min_f \mathcal{L}(f(X), \mathbf{y}) + \alpha \|f\|
\end{gather}
where $\mathcal{L}$ is the loss function, and $\alpha$ is the regularization coefficient.

In Graph Neural Networks (GNNs), the classifier function additionally assumes access to a graph between the data points in $X$. Let this graph be $G$. Then, this classifier function is often estimated by solving,
\begin{gather}
    f^* = \arg \min_f \mathcal{L}(f(X, G), \mathbf{y}) + \alpha \|f\|
\end{gather}

But, often in practice, we may not have access to any graphs. The goal of our work here is to estimate a classifier function by solving,
\begin{gather}
\label{eq:framework}
    f^*, G^* = \arg \min_{f, G} \mathcal{L}(f(X, G), \mathbf{y}) + \alpha \|f\| + \beta \Omega(G, \mathbf{y})
\end{gather}
where $\Omega$ is additional loss terms which could depend on $G$ and the known labels $\mathbf{y}$ and $\beta$ is its coefficient.

In the next section, we will discuss a few related works that address this problem. For all further discussions, we will use the Graph Convolutional Network (GCN) as the Graph Neural Network (GNN) unless otherwise specified.

%% file: sections/related.tex
\section{Related Work}
\label{sec:relatedwork}
In Section~\ref{sec:problem}, we formally described the problem as estimating both graph and the classifier function in the semi-supervised classification setting.

\textbf{IDGL:} The closest related work is Iterative Deep Graph Learning (IDGL)~\cite{idgl}, where we can describe the loss function presented in the paper in the framework of Equation~\ref{eq:framework}. This method assumes access to an initial graph and then iteratively refines it as follows. IDGL starts by training a Graph Convolutional Network (GCN) with a fixed initial graph. A similarity metric function is learned using the latent node representations of trained GCN. The output of the similarity metric is thresholded and combined with the original input graph. This procedure is repeated for multiple iterations. This method works well if the latent node representations learnt by GCN are good; however, when the initial graph is noisy, they are likely to be poor. Thereby, this model gives very small to no improvement when \textit{k}NN graphs are utilized as the initial graph. Our proposed approach mitigates this issue by removing the dependency on learnt representations of GCN.

\textbf{Attention-based:} Attention-based models, Graph Attention Network (GAT)~\cite{gat} and SuperGAT~\cite{supergat}, also fall within the framework of Equation~\ref{eq:framework}. Both these models can be initialized with \textit{k}NN graphs where attention is placed on the edges of this graph. GAT doesn't have any explicit loss on attention itself, while SuperGAT does place a loss on attention, but it is a self-supervised loss. Learning attention on all edges of a graph is difficult given the limited amount of labeled data. In our proposed approach, we restrict our attention to the edges from all the nodes to the labeled set nodes to mitigate this problem. This restriction allows us to utilize the labeled data to its full extent. 

\textbf{LDS:} Learning Discrete Structure (LDS)~\cite{lds} takes an initial graph and proposes to learn a discrete distribution over the edges of the graph. 
Using the validation set, the parameters of this distribution are tuned. Effectively, this paper treats the graph as hyperparameter and tunes it over the validation set. While this paper is about learning graphs, it differs from the framework presented in Equation~\ref{eq:framework}. Equation~\ref{eq:framework} uses training data to estimate both the function and the graph. Model selection is made using the validation set. However, in LDS, the graph itself is tuned over validation. First of all, the method's reliance on the validation set for graph learning can be inhibiting because often validation sets tend to be much smaller than training data. Secondly, this method has a complicated optimization and is very expensive to train. In our work, we mitigate these issues, by proposing a joint learning model along the lines of IDGL, GAT and SuperGAT, and the model is simple and much faster to train. 

Other works include PG-Learn~\cite{pglearn}, which is similar to LDS. It proposes a parameterized graph that can be tuned directly over the validation set. \cite{grcn} is similar to IDGL; it proposes an iterative refinement of the graph based on the latent node representations learnt by the GCN model. However, one can see it as a special case of IDGL where the similarity metric is fixed. \cite{togcn} is graph refinement method based on the assumption that nodes connected in a graph must have the same labels. The refinement update equation obtained updates the graph based on GCN's latent node representations similar to IDGL. In another related work~\cite{gam}, authors propose a Graph Agreement Model (GAM), in which the model learns to predict whether two nodes will have the same label or not. The predictions of this model are utilized as regularization coefficients for the classifier. This work shares some similarities with our Label-Affinity model; however, there are some key problems. As with attention, learning an all pair label agreement model should be difficult and challenging with limited labeled data.  However, the reported results in the paper look very impressive. Upon inspection of their code, we discovered a label leakage bug, and we notified the authors about it. After fixing the label-leakage, the numbers show marginal improvement over baselines. We refrain from comparing with this model due to these issues. We compare against IDGL, LDS, GAT and SuperGAT as the primary baseline methods in our experiments.

Also, there are several papers on generative models for graphs~\cite{netgan, graphopt} that study graph formation. While these models can be extended to graph learning, analysis of these models for graph learning requirements is beyond this work's scope. In the next section, we describe the details of our model.

%% file: sections/model.tex
\section{Proposed Model}
\label{sec:proposedmodel}

\begin{figure*}%
    \centering
    \label{fig:model}
    \includegraphics[width=0.8\textwidth]{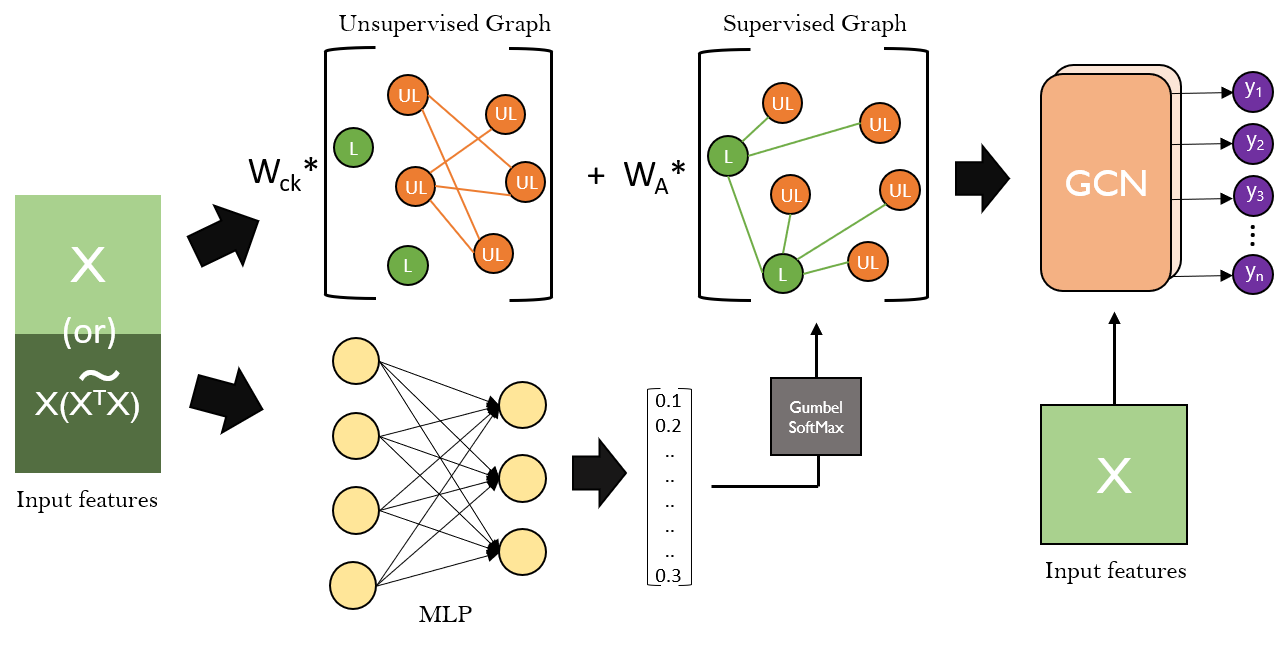}
    \caption{Schematic representation of our approach GLAM}
    \label{fig:glam}
\end{figure*}

The proposed model consists of three basic components: Label-Affinity Model, Convex Combination of Graphs and finally GCN. The architecture diagram is presented in Figure~\ref{fig:glam}

\subsection{Label-Affinity Model}
\label{subsec:labelaffinity}

\begin{figure}%
    \centering
    \subfigure[Cora]{
        \label{fig:noise_cora}
        \includegraphics[width=0.22\textwidth]{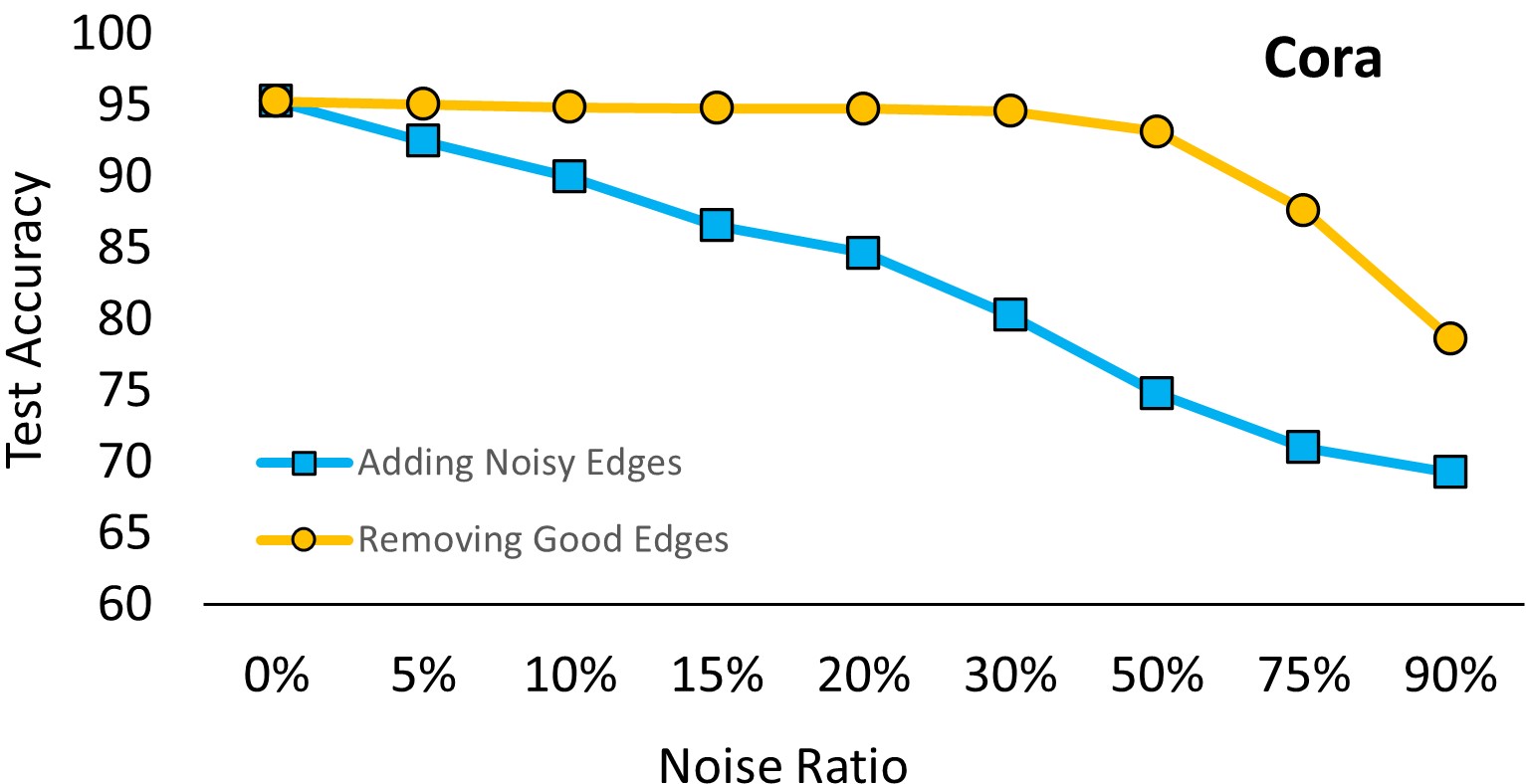}
    }
    \subfigure[Citeseer]{
        \label{fig:noise_citeseer}
        \includegraphics[width=0.22\textwidth]{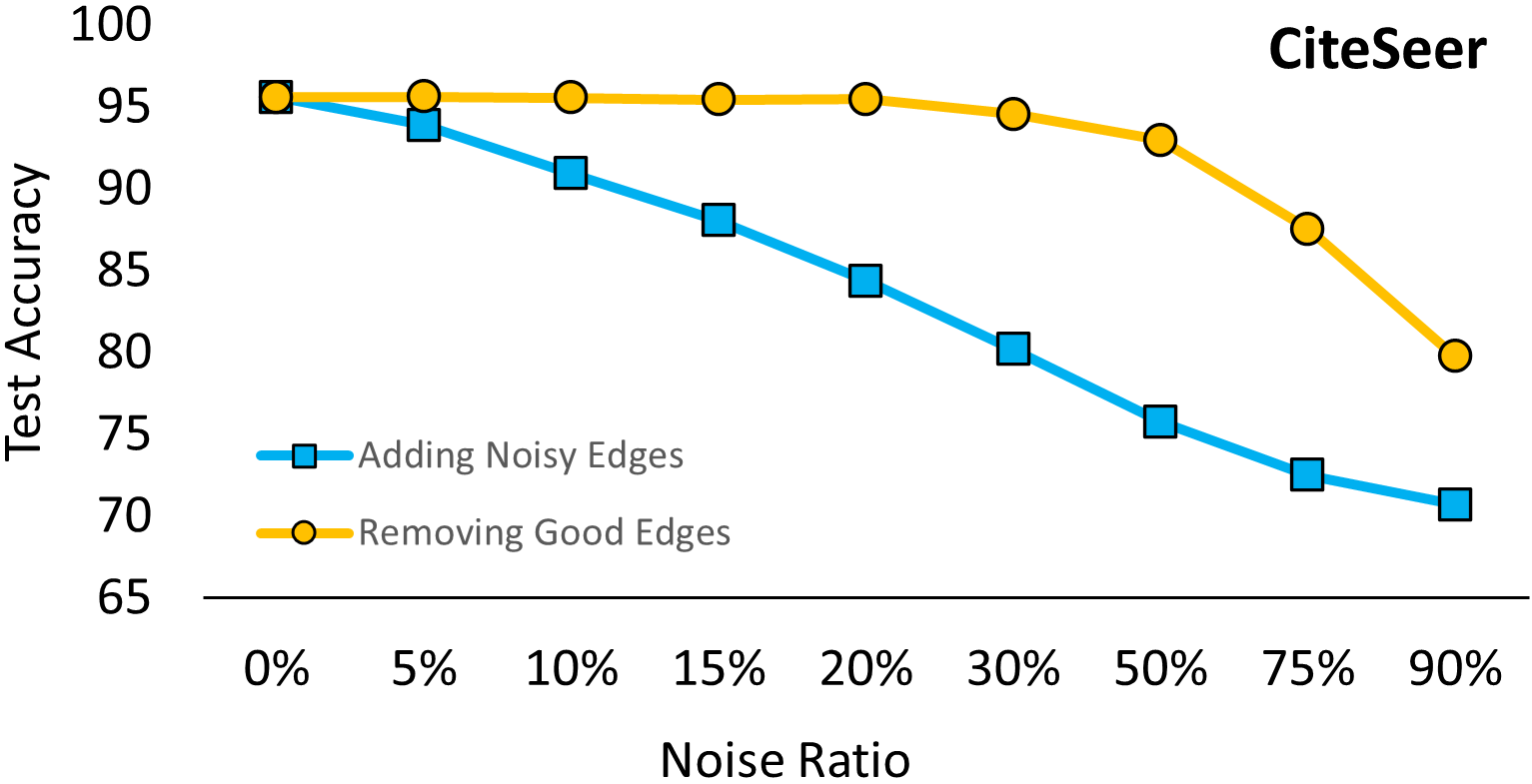}
    }
    \\
    \subfigure[ACM]{
        \label{fig:noise_acm}%
        \includegraphics[width=0.22\textwidth]{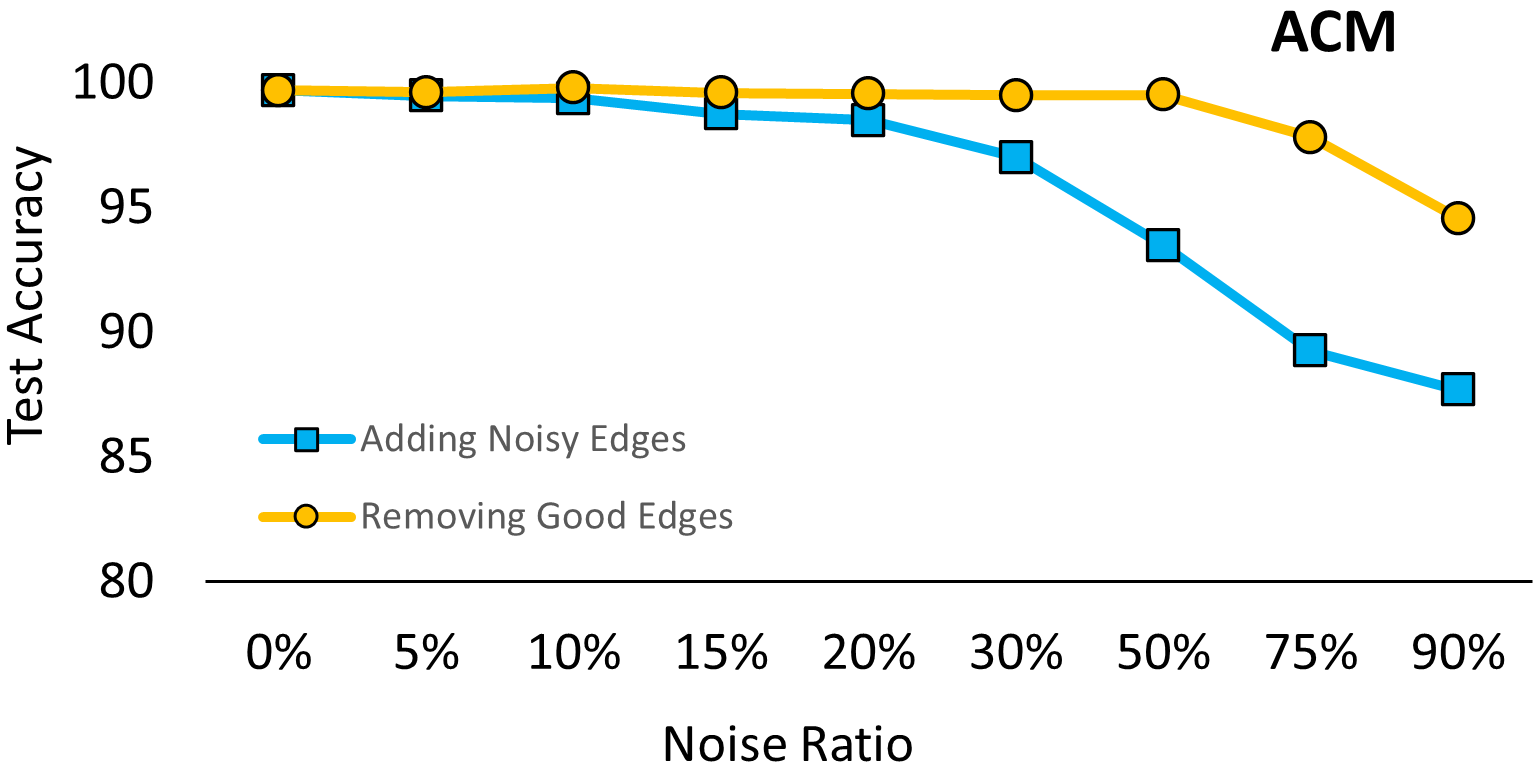}
    }
    \subfigure[DBLP]{
        \label{fig:noise_dblp}%
        \includegraphics[width=0.22\textwidth]{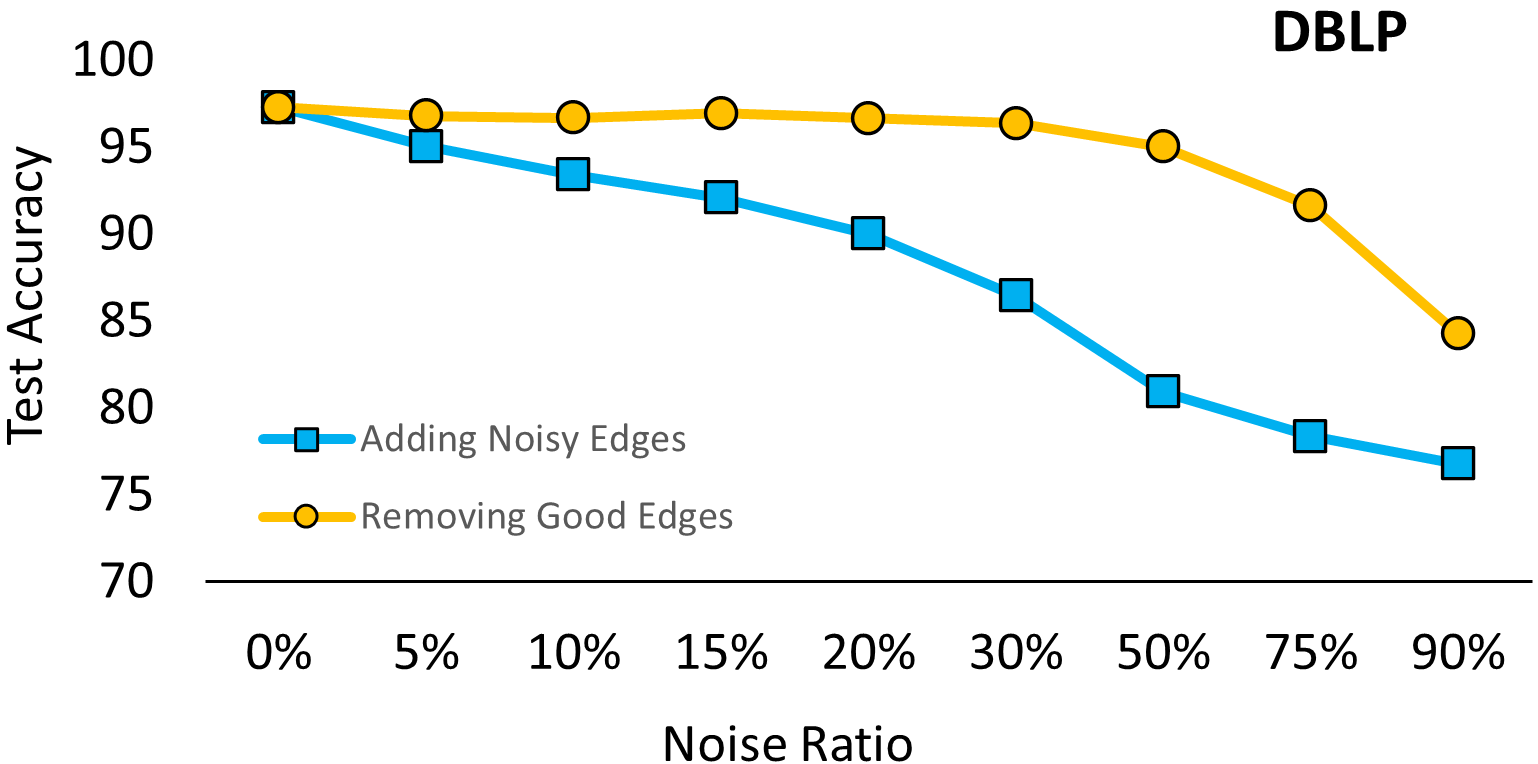}
    }
    \caption{Noise Analysis Plots}
    \label{fig:noiseanalysis}
\end{figure}

Attention-based models attempt to learn attention over all the edges of a graph. Learning these attentions can be a big challenge when working with limited labeled data. The critical insight here is to realise that having a few noisy edges can hurt more than throwing away several good edges (i.e. edges where the source and target nodes have same labels). To illustrate this point, we do a simple analysis. We construct \textit{k}NN graphs for several of the benchmark datasets. We throw away all the noisy edges from this graph and call this as Perfect-\textit{k}NN graph. We conduct two experiments - 1] We randomly add noisy edges to this Perfect-\textit{k}NN graph, and 2] We randomly remove good edges from the Perfect-\textit{k}NN graph. We report test accuracies (using GCN) for varying percentages of added noisy edges and removed good edges. Figure~\ref{fig:noiseanalysis} shows these plots. We observe that adding noisy edges deteriorates the performance of the model. However, even after removing 50-75\% of good edges, the GCN can still perform well on the test set. This observation suggests that it is important to reduce the amount of noise added rather than saving good edges.

Based on the above analysis, it is necessary to pay attention to edges with more confidence. The assumption we then made here is that we will be more confident about predicting edges from any node to labeled node, instead of any arbitrary edge. Conventional attention-based models attempt to learn attention as a function $X \times X \mapsto \mathbb{R}$. However, we decided to change the model to $X \mapsto \Delta^{|D|}$ where $X$ is the set of instances, $\Delta$ is the simplex and $D$ is the set of labeled data as indicated in Section~\ref{sec:problem}. This form of modeling was a straight-forward way of enforcing our restriction. This model predicts a distribution over the labeled set. The probability value indicates how likely it is for the target labeled node to have the same label as the input node. We model this with a simple two-layer neural network as follows:
\begin{gather}
Z^A = \textsc{SoftMax}(\sigma_1(\textbf{X}W_1)W_2)
\label{eq:affinity-model}
\end{gather}
where $\textbf{X} \in \mathbb{R}^{l+u \times d}$ is the feature matrix, $d$ is the number of features, $\sigma_1$ is a non-linear activation function and $W_1, W_2$ are model weights. There are two advantages to this model - 1] it is a simple way of enforcing restriction on edges that we desire, 2] it also allows us to write an explicit loss function using the labeled data (we will discuss this in more detail below).

The label-affinity model can be used to construct the affinity graph. To construct a graph, for every node, we compute its predicted distribution over the labeled set and add the edge with the highest probability. This edge is added in both directions. We can write this mathematically as,
\begin{gather}
G_A = \textsc{One-Hot}(\textsc{Arg-Max}(Z^A)) \label{eq:affinity-neighbor}\\
G_A = [G_A : \mathbf{0}^{l+u \times u}] \label{eq:augmenting-unlabeled}\\
G_A = G_A + G_A^{\top} \label{eq:symmetrizing-graph}
\end{gather}
In Equation~\ref{eq:affinity-neighbor}, we get the highest affinity labeled node for all the nodes. In Equation~\ref{eq:augmenting-unlabeled}, we augment the remaining unlabeled columns which are simply zeroes to complete the full graph. In Equation~\ref{eq:symmetrizing-graph}, we make the graph symmetric. However, since $\textsc{Arg-Max}$ is not differentiable, we use the Gumbel-Softmax trick~\cite{gumbelmax} and generate samples from Gumbel-Softmax distribution instead to create the graph.

Now, relying simply on the classifier loss to learn the parameters would be difficult as the gradients that flow into this model won't be very large. Thereby, we add explicit loss for this model by constructing a labeled data from $D$. For every $x_i \in D$, we create a new label vector $y^A_i$ such that,
\begin{gather}
    y^A_i[j] = \begin{cases} 1 & i \neq j, \;\; y_i = y_j \\ 0 & \textrm{otherwise} \end{cases}
\end{gather}
where $y^A_i[j]$ is the $j^{th}$ element of the vector. Further it is normalized as $y^A_i = y^A_i / \mathbf{1}^{\top}y^A_i$. We use this to define the loss for the affinity model as follows:
\begin{gather}
    \mathcal{L}_A = \sum_{i = 1}^l \textsc{Cross-Entropy}(Z^A_i, y^A_i)
    \label{eq:affinity-loss}
\end{gather}
where $Z^A_i$ is the row corresponding to the $i^{th}$ point in $X$.

In the next section, we will show how the graph constructed from the affinity model is combined with the unsupervised \textit{k}NN graph.

\subsection{Combining Graphs}
\label{subsec:graphcombination}
In Section~\ref{subsec:labelaffinity}, we discussed the construction of affinity graphs. Here, we discuss how we combine it with the unsupervised \textit{k}NN graphs. We need to combine them in the first place because the affinity graph alone is insufficient since there are not many sufficient edges for feature propagation, particularly between unlabeled nodes. Thereby, we utilize the \textit{k}NN graph to assist in that regard. However, before we combine them, we need to pre-process the \textit{k}NN graph. Note that, labeled nodes are our only known ground truth. \textit{k}NN graphs can add a lot of noisy edges and corrupt the feature vectors of the training nodes. To avoid such a possibility, we first crop all the incoming edges to the labeled nodes. We recognize that there is noise in other edges of the graph, but it is difficult to filter this noise. However, removing incoming noise to training nodes gives improvements as we show in Section~\ref{subsec:ablative}. We call this as $G_{ck}$, where $ck$ stands for cropped \textit{k}NN graph. This graph is then combined with the affinity graph as follows:
\begin{gather}
    G = w_AG_A + w_{ck}G_{ck} \; \textrm{s.t.} \; w_A + w_{ck} = 1
    \label{eq:graphcombination}
\end{gather}

This process is similar to the combination of graphs done in IDGL, except that we are not utilizing the latent node representations from GCN, instead we are using both GCN classifier loss and the affinity loss defined in Equation~\ref{eq:affinity-loss} to influence the graph learning.

\subsection{GCN}
\label{subsec:gcn}

In Section~\ref{subsec:graphcombination}, we discussed how the graphs can be combined. This combined graph is finally fed to a two-layer GCN model. However, because of the cropped \textit{k}NN graph, this combined graph is asymmetric. For computing the Laplacian, we can use either the indegree or the outdegree matrix. In our experiments, we have utilized the indegree matrix. Let $D_{in}$ be the indegree matrix of $G + I$ where $I$ is the identity matrix. We can represent the two-layer GCN acting on the combined graph mathematically as follows:
\begin{gather}
    Z^C = \textsc{SoftMax}(\hat{G}\sigma_2(\hat{G}\textbf{X}W_3)W_4)
\end{gather}
where $\hat{G} = D_{in}^{-1/2}(G+I)D_{in}^{-1/2}$, $X \in \mathbb{R}^{l+u \times d}$ is the feature matrix, d is the number of features, $\sigma_2$ is a non-linear activation function and $W_3, W_4$ are model weights.

Finally, we define the classifier loss as,
\begin{gather}
    \mathcal{L}_C = \sum_{i=1}^l \textsc{Cross-Entropy}(Z^C_i, y_i)
    \label{eq:classifier-loss}
\end{gather}
where $Z^C_i$ is the row corresponding to the $i^{th}$ point in X.

So, the final objective that we minimize is,
\begin{gather}
    \mathcal{L} = \mathcal{L}_C + \beta \mathcal{L}_A + \alpha_A \sum_{i = 1}^2 \|W_i\| + \alpha_{C}\sum_{i = 3}^4 \|W_i\|
    \label{eq:complete-loss}
\end{gather}
where $\alpha_A, \alpha_C$ are regularization coefficients. 

The first term in Equation~\ref{eq:complete-loss} refers to the classifier loss, which takes the combination of unsupervised \textit{k}NN and supervised affinity graph (Equation~\ref{eq:graphcombination}). The second term is the affinity loss (Equation~\ref{eq:affinity-loss}) which models label affinity from all nodes to labeled nodes. The affinity loss together with classifier loss improve affinity predictions. Improvement in affinity predictions improves classifier performance. 

\subsection{Note on Input Features}
\label{subsec:inputfeatures}
Note that, for the construction of the \textit{k}NN graphs in Section~\ref{subsec:graphcombination} as well the input to the affinity model in Equation~\ref{eq:affinity-model}, we have used $\textbf{X}$ (the feature matrix). However, in a recent work on text classification, the authors of HeteGCN~\cite{hetegcn} have proposed an architecture which utilizes a normalized version of $\textbf{X}^{\top}\textbf{X}$, call it $\widetilde{\textbf{X}^{\top}\textbf{X}}$ as a first layer of GCN. Essentially they are utilizing feature correlation as the second layer graph. The simplified version of this model is simply equivalent to $\textbf{X}\widetilde{\textbf{X}^{\top}\textbf{X}}$. We believe that the HeteGCN model is benefitting from this feature correlation matrix and that utilizing $\textbf{X}\Tilde{\textbf{X}^{\top}\textbf{X}}$ instead of $\textbf{X}$ should benefit our graph learning problem as well. We simply refer to them as boosted features. We use these boosted features in \textit{k}NN construction and as input to affinity model. We show experimental results with both the normal and the boosted features and not for just our model but for all the baselines, to illustrate how these simple modified features can give significant performance improvements.

%% file: sections/experiment.tex
\section{Experiments}
\label{sec:experiment}
We conducted several experiments to highlight the effectiveness of GLAM against several baselines and state-of-the-art graph learning methods. We restrict our setting to semi-supervised node classification problems where a graph structure is not available. We also present empirical analysis to illustrate why our model performs better.

\subsection{Datasets}
We evaluate on five citation network datasets: Cora, CiteSeer, PubMed taken from \cite{ica}, and ACM, DBLP taken from \cite{han}. In the DBLP dataset, each node represents an author. In the rest datasets, nodes correspond to documents (scientific papers). The node features correspond to sparse bag-of-words features with either binary or TF-IDF values. These datasets are evaluated for node classification task in a transductive setting closely following the experimental setup of \cite{planetoid}. For Cora, CiteSeer, and PubMed datasets, we use the standard split from previous work \cite{gcn}. For ACM and DBLP datasets, we fix the validation and test set to 500 and 1000 nodes and create a training set by sampling 20 nodes per class from the remaining nodes to be in line with the standard split. Detailed statistics of the datasets used are available in the Table~\ref{tab:stats_table}.

\begin{table}[]
\centering
\resizebox{0.45\textwidth}{!}{%
\begin{tabular}{|c|c|c|c|c|c|}
\hline
\textbf{} & \textbf{Cora} & \textbf{Citeseer} & \textbf{Pubmed} & \textbf{ACM} & \textbf{DBLP} \\ \hline
\textbf{Nodes}         & 2,708 & 3,327 & 19,717 & 3,025 & 4,057 \\ \hline
\textbf{Features} & 1,433 & 3,703 & 500 & 1,830 & 334 \\ \hline
\textbf{Classes}     & 7 & 6 & 3 & 3 & 4 \\ \hline
\textbf{No. Training nodes}      & 140 & 120 & 60 & 60 & 80 \\ \hline 
\end{tabular}%
}
\caption{Datasets Statistics}
\label{tab:stats_table}
\end{table}

\subsection{Baselines}
We compare GLAM against several baselines covering simple non-graph based approaches, semi-supervised classification methods, graph neural networks, and state-of-the-art graph learning approaches. Below are the hyper-parameters ranges we followed for tuning the baselines. For all the methods that rely on graphs, we construct \textit{k}NN graphs from input features using cosine metric. Number of neighbors, \textit{k}, is a hyper-parameter and swept over \{5, 10, 15, 20\}. 

\noindent\textbf{LogReg:} Logistic Regression's weight-decay hyper-parameter, C, is tuned over [1e-4, 1e4] in powers of 10 on the validation set.

\noindent\textbf{MLP:} We employ a Multi Layer Perceptron with 1 hidden layer. For tuning, hidden layer dimensions were swept over \{32, 64, 128\}, weight decay over [1e-4, 1e4] in logarithmic steps, learning rate over [1e-3, 1e-2, 1e-1], and dropout from (0, 1).

\noindent\textbf{LP:} \cite{lp_zhu} In Label Propagation, we tune the hyper-parameter $\alpha$ (clamping factor) over the range (0, 1) in steps of 0.01. We observe much better results for LP than reported in LDS because of this extensive tuning.

\noindent\textbf{ManiReg:} \cite{manifold_reg} Manifold regularization's hyper-parameters $\gamma$\textsubscript{A} and $\gamma$\textsubscript{I} are tuned from the range (1e-5, 1e2) in logarithmic steps. We observe a discrepancy in ManiReg's numbers reported in LDS. \cite{gcn} reported 59.5 and 60.1 as test performance on Cora and Citeseer datasets using the original graphs that are part of the datasets. LDS reports 62.3 and 67.7 as mean test accuracy for these datasets using \textit{k}NN graphs. However, \textit{k}NN graphs are of poor quality in terms of homophily and GNN performance on these datasets. 

\noindent\textbf{SemiEmb:} \cite{embednn} Semi-Supervised embedding's hyper-parameters $\lambda$ is tuned over (1e-5, 1e2) in logarithmic steps, hidden layer dims over \{32, 64, 128\}, weight decay over [1e-4, 1e4] in steps of 10, learning rate from \{1e-3, 1e-2, 1e-1\}, and dropout from (0, 1).

\noindent\textbf{GCN:} \cite{gcn} We used 2 layered Graph Convolutional Networks and follow the hyper-parameter ranges mentioned in \cite{pitfalls} for tuning.

\noindent\textbf{GAT}: \cite{gat} For tuning Graph Attention Networks, we consulted \cite{pitfalls} for hyper-parameters ranges.

\noindent\textbf{SuperGAT}: \cite{supergat} we rely on authors\footnote[1]{https://github.com/dongkwan-kim/SuperGAT} code for experiments and follow \cite{supergat} for tuning.

\noindent\textbf{LDS:} \cite{lds} We rely on authors\footnote[2]{https://github.com/lucfra/LDS-GNN} code to perform LDS experiments on our benchmark datasets. We follow the hyper-parameters ranges mentioned by the author.

\noindent\textbf{IDGL}: \cite{idgl} We rely on authors\footnote[3]{https://github.com/hugochan/IDGL} code to perform IDGL experiments on our datasets. For PubMed dataset, we ran IDGL-Anchor variant to report numbers and for the rest datasets, IDGL base variant was used for conducting experiments. The hyper-parameters mentioned by the author are used for tuning the model.

\noindent\textbf{GLAM:} Hyper-parameters $\alpha_A$ and $\alpha_C$ are tuned from (1e-5, 1e4) in logarithmic steps, learning rate from (1e-3, 1e0), dropouts from (0, 1), $W_{ck}$ from (0, 1), affinity classifer's hidden layer dims from \{32, 64, 128, 256\}, GNN's hidden layer dims from \{16, 32, 64, 128\}, \textit{k}NN's \textit{k} from \{5, 10, 15, 20\}, and gumbel softmax's temperature is set to 1e-10. We use Adam optimizer to minimize our combined loss term. GLAM is trained for 500 epochs with an early stopping criterion of no improvement in validation accuracy for 25 epochs. Hyper-parameter tuning was done using \textit{optuna}~\cite{optuna}. We swept through 2000 configurations using TPE sampler in \textit{optuna} for each dataset.

All the models we implemented except LP and LogReg were written in \textit{Tensorflow} \cite{tensorflow}. LP and LogReg were implemented using scikit-learn python package \cite{scikit}. For all models, test accuracy is reported for the configuration that achieves the highest validation accuracy. 

\subsection{Experimental Results}
We report the mean test accuracy and standard deviation over five random seeds for all benchmark datasets in Table~\ref{tab:main_table}. GLAM shows competitive performance across all benchmark datasets except PubMed with accuracy gains of up to 3\%. We further improve all the baselines by feeding them with the boosted features as mentioned in Section~\ref{subsec:inputfeatures} or \textit{k}NN graphs constructed from these boosted features accordingly. These boosted features are computed by multiplying features with feature-feature correlation matrix computed by using outer-product. Our results in the Table~\ref{tab:main_table} shows that these new features, in most cases, greatly improve baseline models with accuracy gains of up to 14\%. Boosted Features part of Table~\ref{tab:main_table} shows that many baselines are competitive and GLAM improves over state-of-the-art approaches by accuracy gains of up to 1.5\%. We discuss results in detail in upcoming subsections.

\begin{table*}[]
\centering
\resizebox{\textwidth}{!}{%
\begin{tabular}{@{}c|ccccc|ccccc@{}}
\toprule
\textbf{} &
  \multicolumn{5}{c|}{$\textbf{Features}$} &
  \multicolumn{5}{c}{$\textbf{Boosted Features}$} \\ \midrule
\textbf{Models\textbackslash{}Datasets} &
  \textbf{Cora} &
  \textbf{CiteSeer} &
  \textbf{Pubmed} &
  \textbf{ACM} &
  \textbf{DBLP} &
  \textbf{Cora} &
  \textbf{CiteSeer} &
  \textbf{Pubmed} &
  \textbf{ACM} &
  \textbf{DBLP} \\ \midrule
\textbf{LogReg} &
  56.20 &
  61.90 &
  73.60 &
  78.80 &
  66.90 &
  70.20 &
  69.90 &
  76.20 &
  88.30 &
  80.20 \\
\textbf{MLP} &
  59.18 (1.39) &
  62.02 (1.80) &
  72.92 (0.55) &
  79.92 (0.55) &
  67.00 (0.53) &
  70.34 (0.56) &
  68.50 (1.52) &
  74.78 (0.89) &
  87.66 (0.69) &
  78.96 (0.66) \\ \midrule
\textbf{LP} &
  54.20 &
  57.40 &
  63.40 &
  75.20 &
  66.30 &
  59.20 &
  54.20 &
  67.70 &
  89.00 &
  65.20 \\
\textbf{ManiReg} &
  58.80 &
  61.10 &
  69.20 &
  80.90 &
  72.60 &
  53.50 &
  55.30 &
  68.10 &
  90.00 &
  73.40 \\
\textbf{SemiReg} &
  66.24 (1.13) &
  66.44 (0.92) &
  73.78 (0.69) &
  88.76 (1.21) &
  74.26 (1.21) &
  69.13 (0.55) &
  70.09 (0.10) &
  74.04 (0.35) &
  88.88 (0.33) &
  79.54 (0.34) \\ \midrule
\textbf{GCN} &
  68.32 (1.37) &
  68.76 (1.04) &
  70.12 (1.05) &
  86.38 (0.53) &
  77.62 (0.95) &
  70.20 (0.81) &
  69.36 (0.67) &
  75.24 (0.39) &
  90.84 (0.39) &
  80.48 (0.34) \\
\textbf{GAT} &
  68.50 (0.99) &
  70.00 (0.76) &
  70.06 (1.83) &
  86.36 (0.26) &
  77.58 (0.81) &
  70.86 (0.63) &
  69.10 (1.35) &
  75.18 (0.39) &
  90.92 (0.34) &
  80.28 (0.49) \\
\textbf{SuperGAT} &
  69.23 (0.31) &
  69.36 (0.58) &
  70.88 (0.68) &
  86.27 (0.31) &
  77.44 (0.44) &
  69.90 (1.67) &
  69.70 (1.29) &
  75.58 (0.25) &
  90.58 (0.93) &
  78.38 (0.95) \\ \midrule
\textbf{LDS} &
  \textbf{70.76 (0.78)} &
  72.16 (0.61) &
  OOM &
  86.98 (0.78) &
  76.72 (0.57) &
  \textbf{72.87 (0.45)} &
  71.44 (0.40) &
  OOM &
  \textbf{92.93 (0.61)} &
  79.42 (0.84) \\
\textbf{IDGL} &
  70.32 (0.54) &
  67.65 (1.71) &
  \textbf{77.20 (0.76)} &
  88.94 (0.52) &
  74.26 (0.84) &
  71.90 (0.69) &
  68.88 (0.44) &
  \textbf{79.04 (0.26)} &
  91.09 (0.38) &
  79.66 (0.47) \\ \midrule
\textbf{GLAM} &
  70.58 (0.18) &
  \textbf{72.22 (0.45)} &
  74.03 (0.32) &
  \textbf{89.34 (0.15)} &
  \textbf{79.70 (0.32)} &
  72.64 (0.35) &
  \textbf{71.86 (0.44)} &
  76.06 (0.35) &
  92.38 (0.20) &
  \textbf{81.52 (0.30)} \\ \bottomrule
\end{tabular}%
}
\vspace{-2mm}
\caption{Mean test accuracy over 5 random seeds for benchmark datasets. Standard deviation is reported in brackets where ever applicable.}
\label{tab:main_table}
\end{table*}

\subsection{What do Attention models capture?}
\label{subsec: vsGAT}
We can view GLAM as an instance of an attention model that places sparse attention on edges between all nodes and labeled nodes. We can now see whether focusing attention on restricted set of edges fares better than learning attention over all the edges of the graph. SuperGAT works under the assumption that if two nodes are linked, they are more relevant to each other than others, and if two nodes are not linked, they are not important to each other \cite{supergat}. The paper suggests that if the homophily of the graph is $> 0.2$, SuperGAT\textsubscript{MX} performs well. However, we observe that GAT performs as well as or better than SuperGAT\textsubscript{MX} with \textit{k}NN graphs. We report homophily percentages for \textit{k}NN graphs on our datasets in Table~\ref{tab:homo_aff_scores}. This phenomenon indicates that there might be more underlying reasons for when attention works or does not. Towards this end, and to quantify the quality of learned attention coefficients, we compute two metrics: bad neighbors ratio and weighted homophily. 

We define bad neighbors for a center node as all the neighboring nodes having different labels than the center node. For Attention-based models, we extract attention matrices (A) from the hidden layers for all heads. For GLAM and GCN, we treat Laplacian matrices as attention matrices. The below equation computes the bad neighbor ratio (BNR):
\vspace{-3mm}
\begin{gather}
  \textsc{BNR} = \frac{1} {|h|} \sum_{h} \frac{  \sum\limits_{i=1}^{|V|} (\mathds{1}_{\textrm{BW}^{(i)} > \textrm{GW}^{(i)}}) } {\sum\limits_{i=1}^{|V|} (\mathds{1}_{\textrm{BW}^{(i)} > 0})}
\end{gather}
where,
    $\textrm{BW}^{(i)} = \sum_{j\in{N_{(i)}}} A_{ij}^{(h)} \mathds{1}_{\ell(i) \neq \ell(j)}$, \\
    $\textrm{GW}^{(i)} = \sum_{j\in{N_{(i)}}} A_{ij}^{(h)} \mathds{1}_{\ell(i) = \ell(j)}$.

In the above equation, the number of attention heads is denoted by h, attention matrix by A, and the total number of nodes by $|V|$. This metric gives a direct insight into how the underlying attention mechanism performs. \\
Weighted homophily is a simple extension of homophily. It is defined as the ratio of attention placed on all good edges to the attention placed on all edges. Tables~\ref{tab:bad_neigh_ratio} and~\ref{tab:weighted_homophily} show these metrics for models in comparison. We find that the bad neighbor ratio aligns well with model performance giving us a perspective on why GLAM performs better.   


\begin{table}[]
\centering
\resizebox{0.45\textwidth}{!}{%
\begin{tabular}{@{}ccccc@{}}
\toprule
\textbf{Weighted Homophily} & \textbf{Cora} & \textbf{CiteSeer} & \textbf{ACM} & \textbf{DBLP} \\ \midrule
\textbf{GCN}      & 61.29 & 62.00 & 86.23 & 73.61 \\\textbf{GAT}      & 58.72 & 60.90 & \textbf{87.21} & \textbf{74.52} \\
\textbf{SuperGAT} & 55.96 & 54.14 & 84.97 & 73.08 \\
\textbf{GLAM}     & \textbf{67.01} & \textbf{69.22} & 85.25 & 73.80 \\
 \bottomrule
\end{tabular}%
}
\vspace{-2mm}
\caption{Weighted Homophily Scores}
\label{tab:weighted_homophily}
\end{table}


\begin{table}[]
\centering
\resizebox{0.49\textwidth}{!}{%
\begin{tabular}{@{}ccccc@{}}
\toprule
\textbf{Bad Neighbor Ratio} & \textbf{Cora} & \textbf{CiteSeer} & \textbf{ACM} & \textbf{DBLP} \\ \midrule
\textbf{GCN}      & 35.95         & 36.32         & 13.32        & 25.98         \\
\textbf{GAT}      & 40.79 (70.00) & 41.33 (67.20) & 14.11 (91.2) & 26.00 (79.60) \\
\textbf{SuperGAT} & 38.98 (71.40) & 39.16 (69.80) & 14.34 (91.1) & 27.21 (78.4)  \\
\textbf{GLAM}     & \textbf{26.05}         & \textbf{26.35}         & \textbf{07.91}        & \textbf{21.43}         \\
 \bottomrule
\end{tabular}%
}
\vspace{-2mm}
\caption{Bad Neighbour Ratio}
\label{tab:bad_neigh_ratio}
\end{table}

\subsection{Can we replace Affinity with Attention?}
Earlier, in Section~\ref{sec:relatedwork}, we argued that attention is restricted to the edges in the graph and thus may not capture long range interactions particularly when the input graph is a \textit{k}NN graph. One might ask, what would happen if we could manually add edges from all nodes to labeled nodes in the cropped \textit{k}NN (Section~\ref{subsec:graphcombination}) and feed it an attention model like GAT? The results for this experiment can be found in Table~\ref{tab:gat_modified}. This suggests that attention models find it difficult to learn good attention coefficients because of limited labeled data.

\begin{table}[]
\centering
\resizebox{0.49\textwidth}{!}{%
\begin{tabular}{ccllll}
\toprule
\textbf{}    & \textbf{Cora} & \textbf{CiteSeer} & \textbf{PubMed} & \textbf{ACM} & \textbf{DBLP} \\ \midrule
\textbf{GAT\textsubscript{ck}} & 39.06 (0.95)  & 35.12 (3.32)      & 56.88 (1.60)    & 68.86 (5.49) & 57.28 (3.29) \\ 
\bottomrule 
\end{tabular}
}
\vspace{-2mm}
\caption{GAT results on modified \textit{k}NN. Standard deviation is reported in brackets.}
\vspace{-2mm}
\label{tab:gat_modified}
\end{table}

\begin{figure*}%
    \vspace{-3mm}
    \centering
    \subfigure[GLAM]{
        \label{fig:tsne_glam}
        \includegraphics[width=0.3\textwidth]{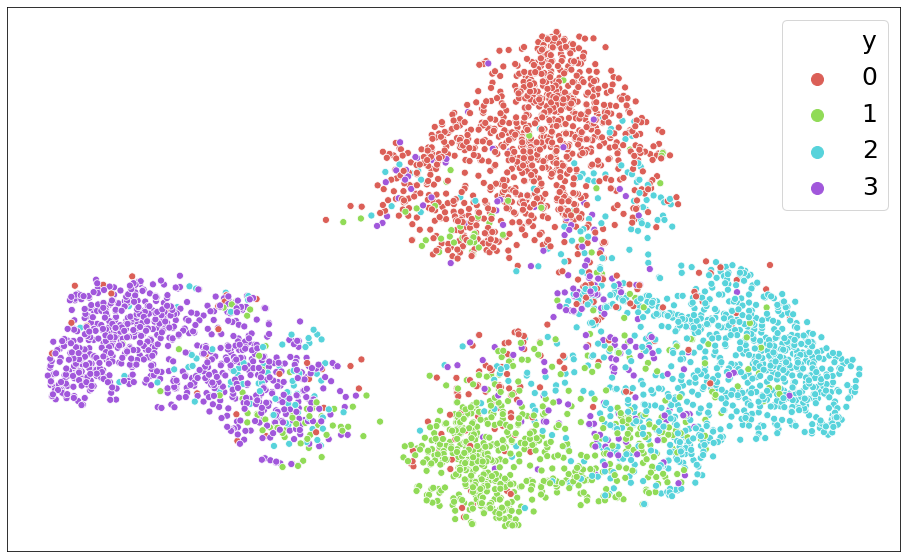}
    }
    \hspace{2mm}
    \subfigure[IDGL - First Iteration]{
        \label{fig:tsne_idgl0}%
        \includegraphics[width=0.3\textwidth]{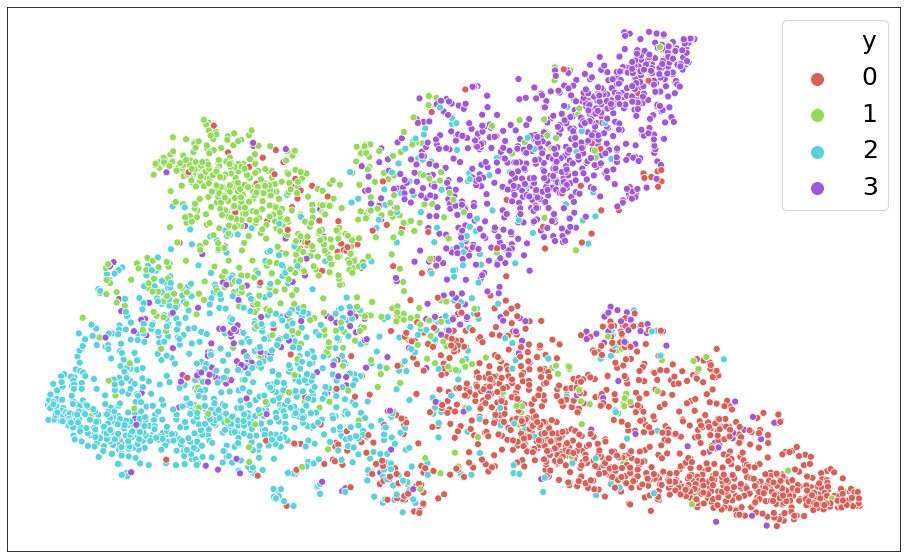}
    } 
    \hspace{2mm}
    \subfigure[IDGL - Last Iteration]{
        \label{fig:tsne_idgl}%
        \includegraphics[width=0.3\textwidth]{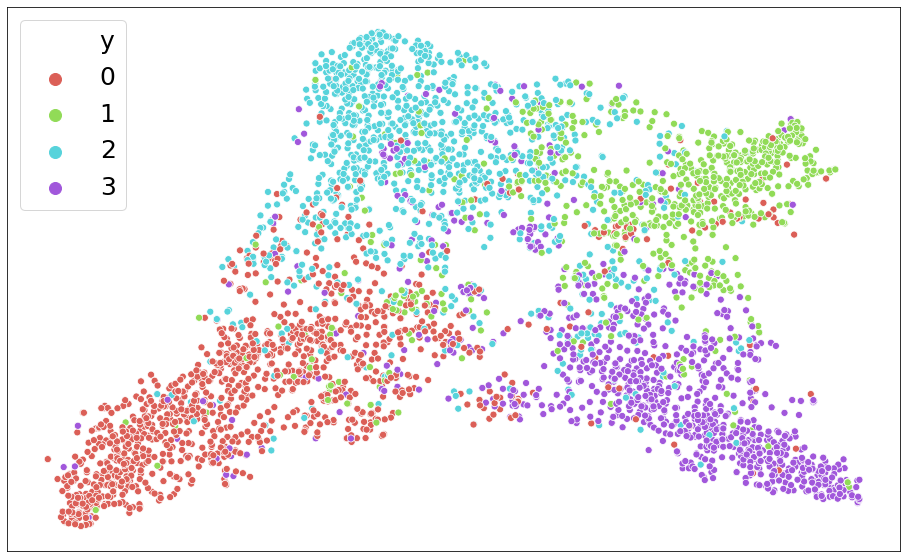}
    }
    \vspace{-2mm}
    \caption{TSNE Plots on DBLP Dataset}
    \label{'tsne_plots'}
    \vspace{-3mm}
\end{figure*}

\subsection{GLAM v/s IDGL: Latent Node Representations}
GLAM and IDGL are similar in the sense that both methods use a convex combination of graphs. IDGL is an iterative approach that jointly learns graph structure and underlying GNN's parameters. IDGL reports significant improvements on citation networks when it uses the graph provided along with the dataset. However, when there is no given graph, IDGL resorts to \textit{k}NN graphs constructed from input features. \textit{k}NN graphs are inferior in quality (homophily) compared to given graphs. For instance, homophily in Cora original graph is 81.00 compared to 58.20 in \textit{k}NN graph. This low-quality graph leads to learning poor representations, and this effect cascades over multiple iterations leading to minor or no accuracy gains. In the case of PubMed, \textit{k}NN graphs quality, 75.08, is close to the graph available along with the PubMed dataset, which has homophily of 79.00. This is the reason IDGL performs better on PubMed compared to GLAM. GLAM works better in the presence of noisy graphs. Figures \ref{fig:tsne_idgl0},~\ref{fig:tsne_idgl} shows IDGL's first and last iteration's embeddings and Figure~\ref{fig:tsne_glam} shows GLAM's embeddings TSNE plots on the DBLP dataset. We observe that the final iteration's plot is similar to the first iteration plot whereas GLAM's plot shows discernible clusters. 

\subsection{Importance of Label Affinity Graph}
We computed the average weight placed on edges in the affinity graph and report it in Table~\ref{tab:homo_aff_scores}. We notice that, in most datasets, W\textsubscript{A} $>$ 0.3 is assigned to affinity edges implying the importance of them. We observe that, in case of the PubMed dataset, large weight is given to \textit{k}NN graph reiterating our point in Section~\ref{subsec: vsGAT} that when \textit{k}NN graph quality is good, they are preferred over affinity graphs.  

\subsection{Is Affinity Graph alone enough?}
\label{sec:affinity-alone}
Table~\ref{tab:affGraphs} shows GCN's performance on using only affinity graph. We see that using affinity graph alone is not sufficient. Affinity graph only contains edges from all nodes to labeled nodes restricting feature propagation, thus limiting the performance of GNNs.

\begin{table}[]
\resizebox{0.48\textwidth}{!}{%
\begin{tabular}{|c|c|c|c|c|}
\hline
\textbf{}                            & \textbf{Cora} & \textbf{CiteSeer} & \textbf{ACM} & \textbf{DBLP} \\ \hline
\textbf{Affinity Graph Weight = 1.0} & 63.88         & 62.8              & 85.52        & 74.43         \\ \hline
\end{tabular}
}
\caption{GCN's performance on Affinity Graph}
\label{tab:affGraphs}
\end{table}

\subsection{Effect of Weight on Affinity Graph}
In this section, we discuss how placing different weights on the affinity graph during graph combination affects GCN's \cite{gcn} performance. We observe that in all the datasets, adding affinity graphs improves the performance of GNNs. Figure~\ref{fig:wteffect} illustrates the effect of the affinity graph on four different benchmark datasets. A bell shape sort of behavior is observed in most cases, where the performance of GNN starts to dwindle as more weight is given to the affinity graph. \textit{K}NN graphs are responsible for feature propagation. Placing more weight on the affinity graph affects feature propagation (as discussion in Section~\ref{sec:affinity-alone}) and we start losing performance after a certain point. 

{
\begin{figure}%
    \centering
    \subfigure[Cora]{
        \label{fig:wt_cora}
        \includegraphics[width=0.22\textwidth]{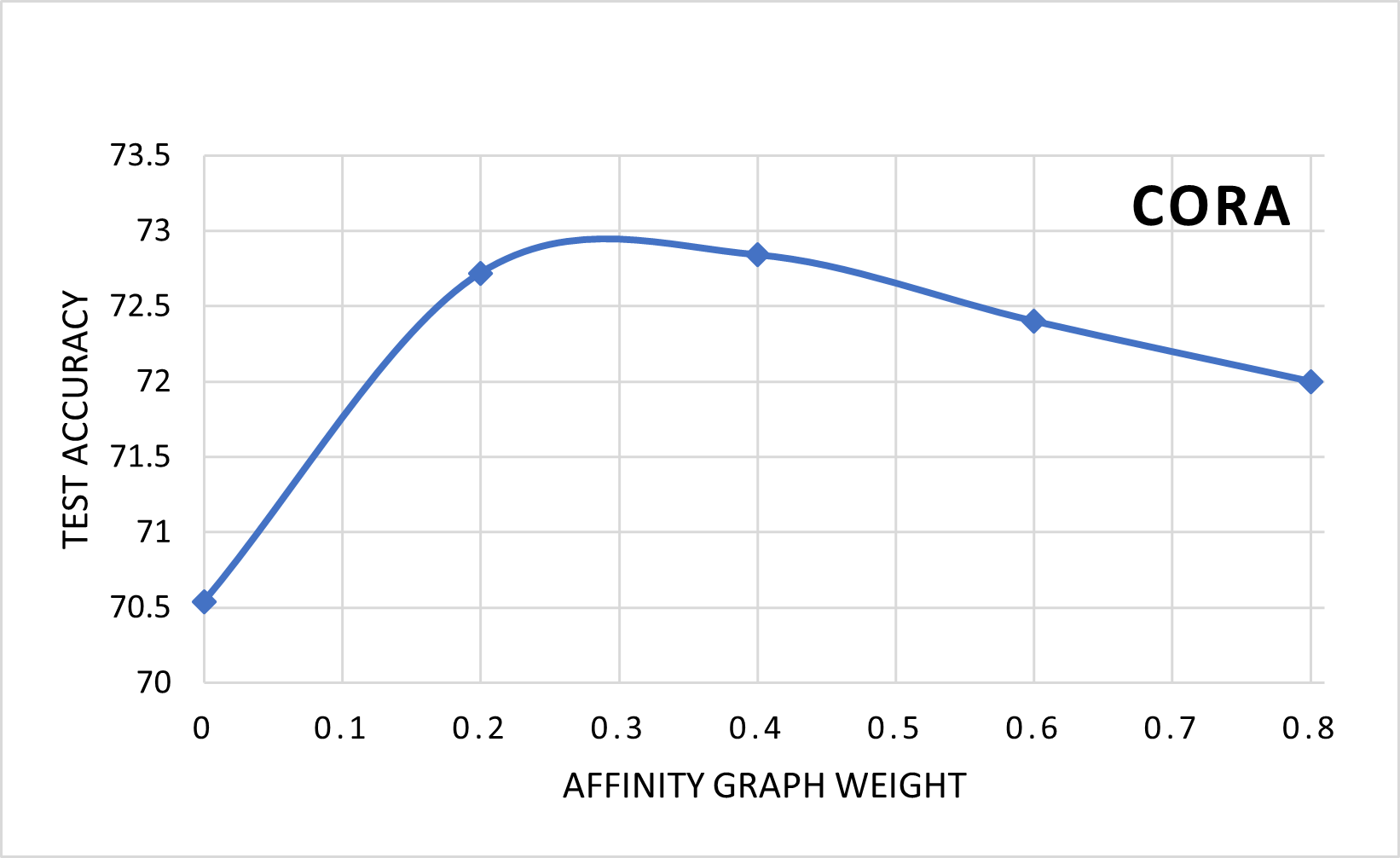}
    }
    \subfigure[Citeseer]{
        \label{fig:wt_citeseer}
        \includegraphics[width=0.22\textwidth]{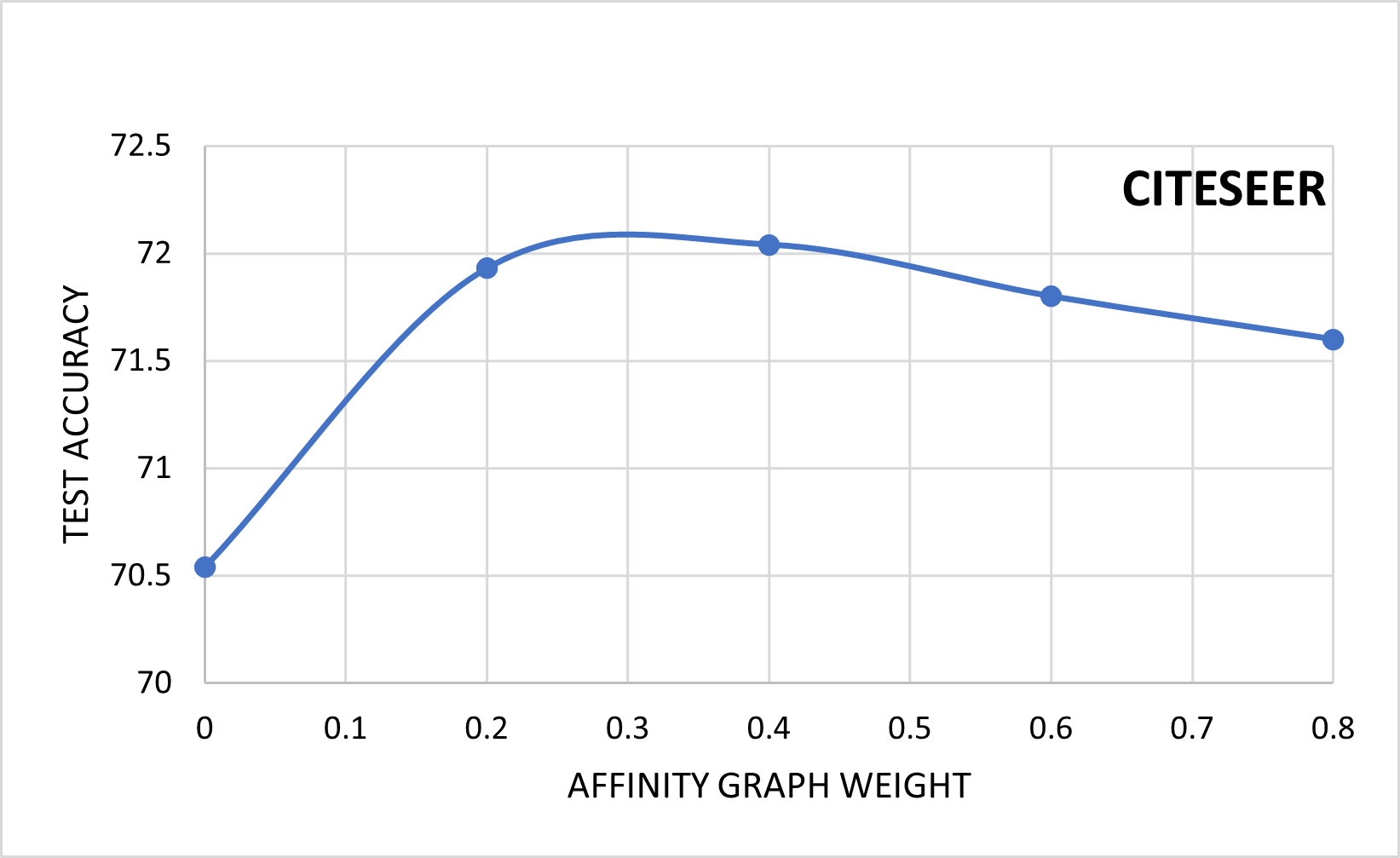}
    }
    \\
    \subfigure[ACM]{
        \label{fig:wt_acm}%
        \includegraphics[width=0.22\textwidth]{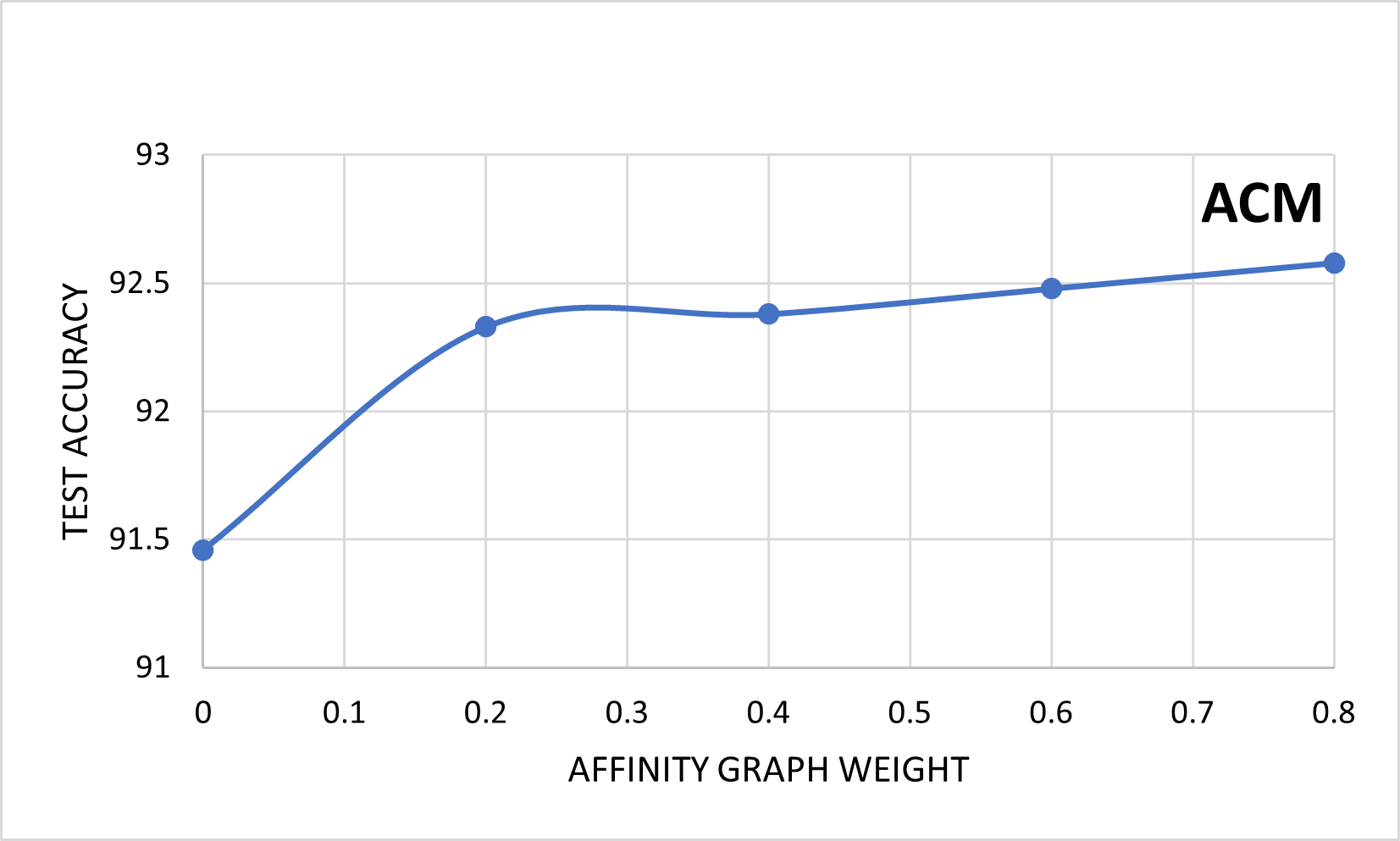}
    }
    \subfigure[DBLP]{
        \label{fig:wt_dblp}%
        \includegraphics[width=0.22\textwidth]{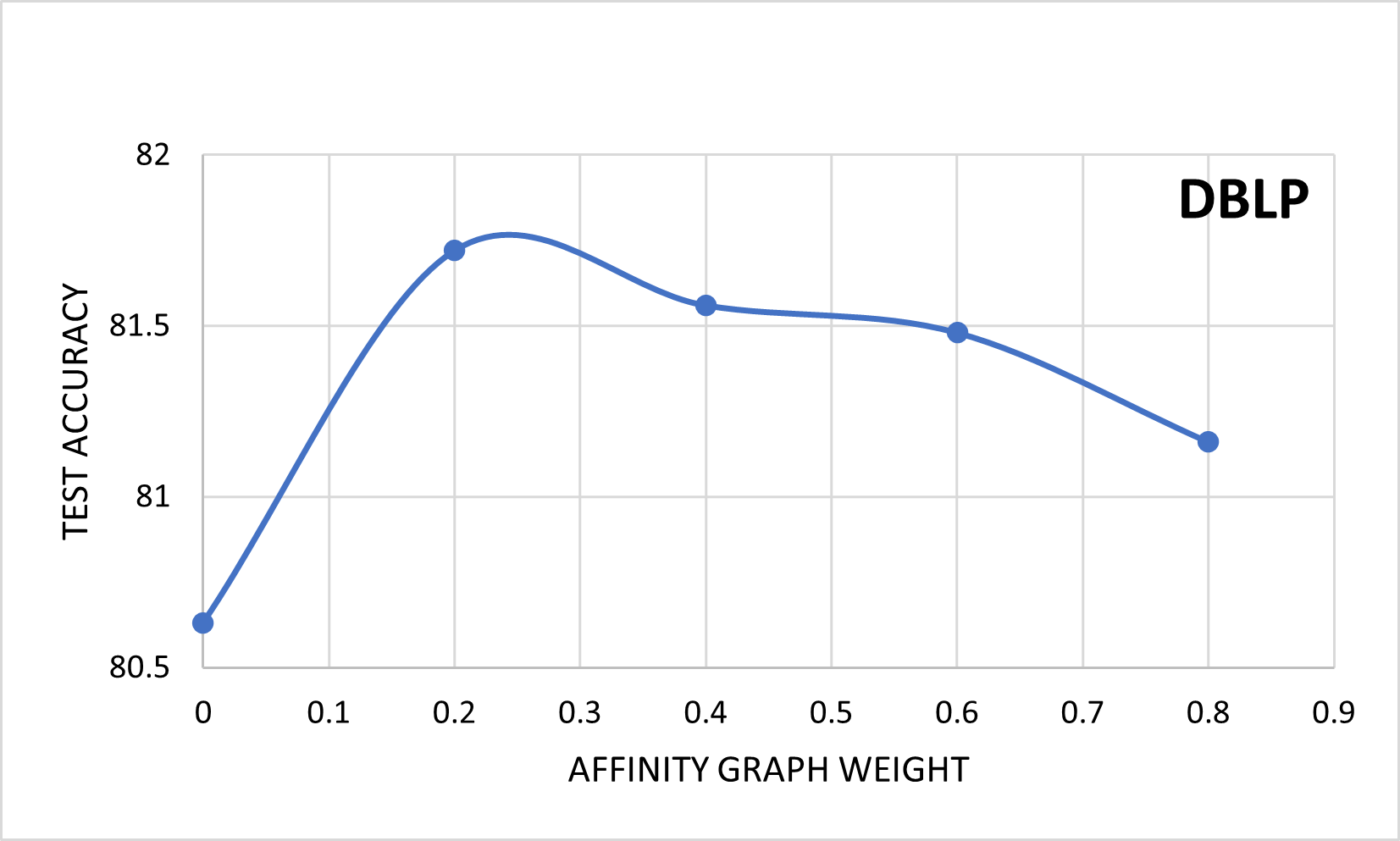}
    }
    \caption{Effect of Weight on Affinity Graph Plots}
    \label{fig:wteffect}
\end{figure}

}

\subsection{Timing Analysis}
Training comparison is done by computing the average end-to-end training time over 2000 runs. All the models in comparison were run on the same machine with Intel Xeon 2.60Ghz processor, 112GB ram, Nvidia Tesla P-100 GPU, Ubuntu 18.04 OS. Table~\ref{tab:timing} shows timing analysis for different benchmark datasets. We observe that we get up to $\sim42\times$ speedup compared to IDGL and up to $\sim70\times$ speed up compared to LDS.

\subsection{Ablation Study} 
\label{subsec:ablative}
Table ~\ref{tab:abl-table} shows the ablative study results of different components in our model. We can see that turning off affinity edges and removing affinity loss significantly affects the performance indicating their importance. It can be observed that performance of GLAM with only cropped \textit{k}NN (w/o affinity graph in Table~\ref{tab:abl-table}) is equivalent to GCN with \textit{k}NN (Table~\ref{tab:main_table}). This suggests that removing all incoming edges to the training nodes does not affect and sometimes improves the performance. This is inline with the noise analysis we presented in Section~\ref{subsec:labelaffinity}. Also, Table~\ref{tab:abl-table} suggests that unless there is an explicit loss for affinity term, we see minuscule improvements over plain \textit{k}NN-GCN model (Table~\ref{tab:main_table}).

\subsection{Preliminary Results using Generative Models}
Graph generative models learn a generator to capture the underlying distribution of the input graph. Graphs are sampled using this generator, making generative models relevant in solving graph learning problems. Multiple graph generation modeling approaches like NetGAN~\cite{netgan} and GraphOpt~\cite{graphopt} have been proposed like mentioned in our paper. In this section, we present preliminary results of NetGAN on the Cora dataset. We construct a \textit{k}NN graph using boosted features and feed it as input to NetGAN~\cite{netgan}. NetGAN learns a generator, G, using the input graph. Following the paper's suggestion, we generate 10,000 random walks of length T (a hyperparameter). We build a Score Matrix, M, where every cell in M indicates edge count between node pairs observed in the random walks. Score Matrix, M, is then normalized, and sample a graph with 'k' edges. This graph fed to a GNN. We observe 70.54 (0.30) mean test accuracy on the Cora dataset; the number in brackets indicates the standard deviation. We see that NetGAN does not show much improvement beyond what \textit{k}NN graphs already offer (\textit{k}NN GCN has a mean test accuracy of 70.20 (0.81) on the Cora dataset). NetGAN is an unsupervised approach that does not rely on available labels to learn the generator. Thus, it limits the performance of these sampled graphs to the input graph's performance. Also, training generative models is a time-taking process and requires a lot of computing power. Graph generative models for graph learning is likely a non-trivial problem. It is possible to combine the Affinity model with these approaches. However, this is beyond the scope of this work.

\begin{table}[]
\centering
\resizebox{0.45\textwidth}{!}{%
\begin{tabular}{lcccc}
\toprule
\multicolumn{1}{c}{\textbf{}} & \textbf{Cora} & \textbf{CiteSeer} & \textbf{ACM} & \textbf{DBLP} \\ \midrule
\textbf{GLAM}                 & \textbf{72.64 (0.35)}  & \textbf{71.86 (0.44)}      & \textbf{92.38 (0.20)} & \textbf{81.52 (0.30)}  \\
\textbf{w/o affinity graph}   & 70.54 (0.76)  & 70.54 (0.76)      & 91.46 (0.42) & 80.63 (0.85)  \\
\textbf{w/o affinity loss}    & 70.90 (0.83)  & 69.06 (1.66)      & 91.46 (0.48) & 80.40 (1.17)  \\ \bottomrule
\end{tabular}
}
\vspace{-2mm}
\caption{Ablation study on various datasets.}
\label{tab:abl-table}
\vspace{-1mm}
\end{table}

\begin{table}[]
\centering
\resizebox{0.475\textwidth}{!}{%
\begin{tabular}{@{}cccccc@{}}
\toprule
\textbf{Average Time} & \textbf{Cora} & \textbf{CiteSeer} & \textbf{PubMed} & \textbf{ACM} & \textbf{DBLP} \\ \midrule
\textbf{IDGL} & 152.68s & 394.94s & 546.81s & 329.64s & 88.89s  \\
\textbf{LDS}  & 327.75s & 683.43s & NA      & 338.25s & 348.75s \\
\textbf{GLAM} & \textbf{5.32s}   & \textbf{22.22s}  & \textbf{23.36s}  &\textbf{7.76s}   & \textbf{5.04s}   \\ \bottomrule
\end{tabular}%
}
\vspace{-2mm}
\caption{Running Time Comparison in seconds}
\vspace{-2mm}
\label{tab:timing}
\end{table}

\begin{table}[ht]
\centering
\resizebox{0.475\textwidth}{!}{%
\begin{tabular}{@{}cccccc@{}}
\toprule
\textbf{Dataset}   & \textbf{Cora} & \textbf{CiteSeer} & \textbf{PubMed} & \textbf{ACM} & \textbf{DBLP} \\ \midrule
\textbf{Homophily} & 58.20         & 59.46             & 75.08           & 87.32        & 85.12         \\
\textbf{W\textsubscript{A}}     & 0.33          & 0.49              & 0.04            & 0.33         & 0.30          \\ \bottomrule
\end{tabular}%
}
\vspace{-2mm}
\caption{Homophilly and average of chosen affinity weights}
\label{tab:homo_aff_scores}
\vspace{-3mm}
\end{table}

%% file: sections/conclusion.tex
\section{Conclusion}
\label{sec:conclusion}

In this paper, we proposed a model to jointly learn graph and classifier for semi-supervised classification tasks where no graphs are available. Our experimental results suggest that our model is fast to train and particularly effective when the unsupervised \textit{k}NN graph is noisy. We analysed and compared with baselines along several dimensions highlighting their limitations and how our model addresses them.


For future work, we want to explore combining the merits of IDGL with our work and see if a single model can work across all possible graphs with different noise levels. Also of great interest would be to combine the proposed idea with some of the graph generative models~\cite{netgan, graphopt}. 

\clearpage
